\documentclass[
]{ceurart}

\usepackage{tcolorbox}
\usepackage{graphicx}
\usepackage{multirow}
\usepackage[normalem]{ulem}
\usepackage{subcaption}
\useunder{\uline}{\ul}{}

\begin{document}

\copyrightyear{2020}
\copyrightclause{Copyright for this paper by its authors.
  Use permitted under Creative Commons License Attribution 4.0
  International (CC BY 4.0).}

\newtcolorbox{mybox}[1]{%
    colbacktitle=gray!30,
    coltitle=black,
    fonttitle=\bfseries,
    colback=gray!30,
    colframe=gray!10,
    sharp corners,
    title=#1}

\conference{HASOC (2021) Hate Speech and Offensive Content Identification in English and Indo-Aryan Languages}

\title{Multilingual Hate Speech and Offensive Content Detection using Modified Cross-entropy Loss}

\author[1]{Arka Mitra}[%
orcid=0000-0003-1071-7294,
email=thearkamitra@gmail.com,
url=https://thearkamitra.github.io/,
]
\address[1]{Indian Institute of Technology, Kharagpur, India}

\author[2]{Priyanshu Sankhala}[%
orcid=0000-0003-2796-0039,
email=priyanshu.nitrr.ele@gmail.com,
url=https://priyanshusankhala.github.io/,
]
\address[2]{National Institute of Technology Raipur, India}

\begin{abstract}
  The number of increased social media users has led to a lot of people misusing these platforms to spread offensive content and use hate speech. Manual tracking the vast amount of posts is impractical so it is necessary to devise automated methods to identify them quickly. Large language models are trained on a lot of data and they also make use of contextual embeddings. We fine-tune the large language models to help in our task. The data is also quite unbalanced; so we used a modified cross-entropy loss to tackle the issue. We observed that using a model which is fine-tuned in hindi corpora performs better. Our team (HNLP) achieved the macro F1-scores of 0.808, 0.639 in English Subtask A and English Subtask B respectively. For Hindi Subtask A, Hindi Subtask B our team achieved macro F1-scores of 0.737, 0.443 respectively in HASOC 2021.
\end{abstract}


\maketitle

\section{Introduction}
With the increased use of social media platform like Twitter, Facebook, Instagram, and YouTube by users around the world, the platforms have had positive aspects including but not limited to social interaction, meeting like-minded people, giving a voice to each individual to share their opinions \cite{Istaiteh2020RacistAS}. However, as a result, social media platforms can also be used to spread hate comments, hate posts by certain individuals or groups; which can lead to having anxiety, mental illness and severe stress to people who consume that hate content \cite{Kawate2017AnalysisOF}. It becomes necessary to be able to detect such activities at its earliest to stop it from spreading, thereby making social media a healthy place to interact and share their views without a fear of getting hate comments\cite{a2019}. \\
The hate posts can be insults or racist or discriminating on the bases of a particular gender, religion, nationality, age bracket, ethnicity. Such comments can also lead to goading of violence amongst people. With the large number of posts being shared each minute, it is not possible to manually classify each of the posts. Thus, a pre-programmed system is required to distinguish Hate speech activities quickly as hate content gains a lot of attention and is subject to be shared fast as well \cite{Mathew2019SpreadOH}. Direct targeted abuses and profane content are not that difficult to classify. However, it is extremely hard to recognize indirect hate content often involving use of humour, irony, sarcasm even for an human annotator when the context of the posts are not provided. This makes the classification task additionally more difficult for most progressive frameworks.\\
HASOC 2021 \cite{hasoc2021mergeoverview} is a shared task for the identification of hateful and offensive content in English and Indo-Aryan Languages. We participated in two sub-tasks for English and Hindi language \cite{cosah}.\\ 
The sub task A refers to classifying twitter samples into: 
\begin{itemize}
    \item \verb|HOF| Hate and offensive :- contains hate speech/profane/offensive content.
    \item \verb|NOT| Non Hate-offensive :- which does not contain any hate speech, profane, offensive content.
\end{itemize}
The sub task B refers to classifying twitter samples into: 
\begin{itemize}
    \item \verb|HATE| Hate speech :- Posts under this class contain Hate speech content.
    \item \verb|OFFN| Offensive :- Posts under this class contain offensive content.
    \item \verb|PRFN| Profane :- These posts contain profane words.
    \item \verb|NONE| Non-Hate :- These posts do not contain any hate speech content.
\end{itemize}
For tasks pertaining to English language, we experimented with large language models like fine-tuning BERT (Bidirectional Encoder Representation from Transformer) \cite{Devlin2019BERTPO}, RoBERTa (A Robustly Optimized BERT Pretraining Approach) \cite{Liu2019RoBERTaAR} and XLNet (Generalized Autoregressive Pretraining for Language Understanding) \cite{Yang2019XLNetGA} out of which RoBERTa outperformed others with the macro F1-score of 0.8089 while BERT and XLNet had the macro F1-score of 0.8050 and 0.7757 respectively in Subtask A and for Subtask B the macro F1-score was 0.6396 with RoBERTa model respectively. 
For the tasks referring to Hindi language, the authors used a model which is fine-tuned on detecting Hinglish sentiments \cite{Lee05} and had the macro F1-score of 0.7379 for Subtask A and macro F1-score of 0.4431 for Subtask B. 
\section{Related Work}

In this section, we will discuss the previous state of the art methods proposed for detection of hate speech. 
The use of BERT and other transfer learning algorithms, and deep neural models based on LSTMs and CNNs tend to perform similar but better than traditional classifiers such as SVM \cite{Modha2020TrackingHI}. The number of papers,  trying to automate Hate speech detection, that have been published in Web of Science has been increasing exponentially \cite{Paz}. Waseem et al. \cite{waseem-etal-2017-understanding} have classified hate speech into different categories and led to the Offensive Language Identification Dataset (OLID) \cite{zampieri-etal-2019-predicting}.\\
There has been work in different sub fields of abuse like in sexism \cite{waseem-hovy-2016-hateful, Jaki2019OnlineHO}, cyberbullying \cite{cyberbullying}, trolling \cite{ws-2018-trolling} and so on. There are hate comments in most of the social media sites like Youtube \cite{Dinakar2011ModelingTD}, Instagram \cite{Zhong2016ContentDrivenDO} which shows the importance of having a generalized Hate detection model \cite{waseem-etal-2017-understanding}. Work done by Yin et al. \cite{YinWenjie2021Tghs} gives an overall idea of the generalizability of the different models that are present for hate speech detection. For the different models, the features from the input that are used have a great impact on the performance. Xu et al. \cite{xu-etal-2012-learning} showed that part-of-speech tags are quite successful for improving the model; it is further improved by considering the sentiment values \cite{Davidson2017AutomatedHS}. The sentences in the online platforms do not always follow the normal textual formats or correct spellings. Thus, Mehdad et al. \cite{mehdad-tetreault-2016-characters} used a character level encoding rather than using the word level encoding proposed by Meyer et al. \cite{meyer-gamback-2019-platform}. The type of architecture used also impacts on the performance on the model. Swamy et al. \cite{swamy-etal-2019-studying} performed a comprehensive study that shows how different models perform and generalize.  

\section{Methodology}

HASOC 2021 \cite{cosah} has been going on for two years now and a lot of different ways are uncovered to detect hate content \cite{Mandl2019OverviewOT, Mandl2020OverviewOT}. This paper covers the use of large language models for classification of hate speech content.
\subsection{Languages}
The Hate speech and Offensive Content Identification in English and Indo-Aryan Languages HASOC 2021 \cite{hasoc2021mergeoverview, cosah} purposes two different tasks, in 3 different languages English, Hindi, Marathi. The authors participated in both tasks for English and Hindi languages.

\subsection{Task description}
The first task in all languages know as "Subtask A" refers to a classification problem of twitter samples which were labelled as \verb|HOF|- Hate and offensive content and \verb|NOT|- Not hate and offensive content.
The second task, know as "Subtask B" refers to a classification of twitter samples which were labelled as \verb|PRFN|- Profane Words, \verb|HATE|- Hate speech, \verb|OFFN|- Offensive Content, and \verb|NONE|- Non-hate content. The detailed description of all columns present in a dataset is given in Table \ref{tab:data_desc} and the number of twitter samples corresponding to each label is given in Table \ref{tab:class_dist}.

\begin{table*}
    \caption{The detailed data description is given in table below:-}
    \label{tab:data_desc}
    \begin{tabular}{|c|c|}
    \hline Columns & Description \\
    \hline tweet\_id & unique value for the tweets \\
    \hline text & full text of the tweets \\
    \hline task1 & label, either tweet is HOF or NOT for Subtask A \\
    \hline task2 & label, either tweet is HATE, OFFN or PRFN for Subtask B \\
    \hline ID & unique hasoc ID for each tweet for Hindi data set\\
    \hline
\end{tabular}
\end{table*}
\begin{table}
  \caption{Class division of both subtasks for Train and Test Dataset}
  \label{tab:class_dist}
    \begin{tabular}{|c|c|c|c|c|c|c|c|}
    \hline \multirow{2}{*} { Subtasks } & \multirow{2}{*} { No. of posts } & \multicolumn{2}{|c|} { Train set } & \multicolumn{2}{c|} { Test set } \\
    \cline { 3 - 6 } & & English & Hindi & English & Hindi \\
    \hline \multirow{2}{*} { Subtask A } & HOF & 2501 & 1433 & 798 & 1027 \\
    \cline { 2 - 6 } & NOT & 1342  & 3161 & 483 & 505 \\
    \hline \multirow{3}{*} { Subtask B } & PRFN & 1196 & 213 & 224 & 74 \\
    \cline { 2 - 6 } & HATE & 683 & 566 & 379 & 215 \\
    \cline { 2 - 6 } & OFFN & 622 & 654 & 95 & 216 \\
    \cline { 2 - 6 } & NONE & 1342 & 3161 & 483 & 1027 \\
    \hline \multicolumn{2}{|c|} { Total } & 3843 & 4594 & 1281 & 1532 \\
    \hline
\end{tabular}
\end{table}

\subsection{Approach}
\begin{figure}[!h]
\begin{minipage}[b]{0.45\linewidth}
\centering
\includegraphics[width=0.5\textwidth]{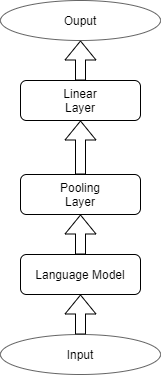}
\caption{Overall Pipeline}
\label{fig:figure1}
\end{minipage}
\hspace{0.5cm}
\begin{minipage}[b]{0.45\linewidth}
\centering
\includegraphics[width=0.6\textwidth]{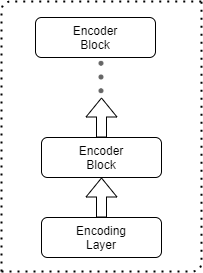}
\caption{Language Model}
\label{fig:figure2}
\end{minipage}
\end{figure}
The dataset that is provided in all the subtasks has an unequal number of samples per class. Table. \ref{tab:class_dist} shows the overall distribution. For subtask A for English, the ratio of the classes (HOF and NOT) is around 2:1 while for Hindi it is around 1:2. Again, for subtask B, the ratio of the classes (PRFN, HATE, OFFN, NONE) is about 2:1:1:2 for english and approximately 2:5:6:30 for Hindi. From the ratio, one can understand that it would be unjust for the loss for each class to be the same. The cross entropy loss assigns same value to a probability score irrespective of the number of times it is present in the training set. To mitigate this, the authors have used modified cross-entropy loss as shown in Eqn. \ref{eqn:modloss}; it assigns a greater loss whenever a class with smaller frequency is misclassified. The weights factor in Eqn. \ref{eqn:modloss} has a higher value for a class if the class has a lower frequency. This penalizes the model whenever that class is wrongly predicted and helps to improve the performance of the model.\\

\begin{equation}
\label{eqn:modloss}
    loss(logits,class) = weight[class]*(-logits[class]+log(\sum_j exp(logits[j])))
\end{equation}
The authors used large-language models since the models are trained on a large amount of data and thus can understand the semantic structure of sentences and the tokens that are sent as inputs to these models have a contextual embedding associated with them. The output of the model is taken and then pooled. The resulting output is then passed through a linear layer and a argmax is used to find the expected class of the sentence as shown in Figure. \ref{fig:figure1}.\\

\section{Results}

The authors submitted four groups of results Table \ref{tab:class_dist_result} gives the final results for our submission. The results has been evaluated on a test dataset, which is about one-third of the training data size, using the Macro F1 scores.

\begin{table}[!h]
  \caption{Results from the official Test set from the leaderboard published from  $15 \%$ the data set}
  \label{tab:class_dist_result}
    \begin{tabular}{cccc}
    \hline Task & Our Score (Macro average F1) & Best Score & Rank \\
    \hline English Subtask A & $0.8089$ & $0.8177$ & 4 \\
    English Subtask B & $0.6396$ & $0.6657$ & 6 \\
    Hindi Subtask A & $0.7379$ & $0.7825$ & 22 \\
    Hindi Subtask B & $0.4431$ & $0.5603$ & 16 \\
    \hline
\end{tabular}
\end{table}
The experiments showed that large cased BERT performed the best followed by RoBERTa and the lowest scores were obtained from the BERT base model. The maximum sequence length that is used has a direct impact on the performance; with a larger length having a better performance, with the training time increases at the same time.\\
The methodology followed for both English and Hindi are the same, but the performance obtained for the English subtask is quite better than that for the Hindi subtask. This shows that the language models are pretty good in understanding the semantics for English but fail to do so for a low resource language like Hindi. The modified cross-entropy loss provided a better F1 score as compared to training with equal importance given to all of the separate classes.

\subsection{Experimental Details}
For English language we experimented with RoBERTa base pre-trained model \cite{Liu2019RoBERTaAR}, fine tuned BERT large cased architecture\cite{Devlin2019BERTPO}, and XLNet  \cite{Yang2019XLNetGA}- all for the same configuration, i.e, max length is set to 120, batch size to 8 and trained with 4 number of epochs. AdamW optimizer \cite{Loshchilov2019DecoupledWD} with an initial learning rate of 2e-5 is used for training. 
Similarly for hindi language tasks we used a pre-trained model \cite{Lee05} from the Hugging face \cite{wolf2020huggingfaces} library. The Max length has been set to 200, batch size was 8, and number of epochs was set to 4.\\
There is a trade-off between the accuracy and the total number of tokens. The amount of time the model takes for training is proportional to the square of the number of tokens. As the number of tokens increases, the amount of time increases. However, when we truncate the maximum length, some of the information present in the sentence gets lost and the prediction for the sentence might be wrong. We had to consider a trade-poff between the accuracy and the time it takes for the model to train. 
For deciding the maximum sentence length, about 99\% percentile of number of tokens in sentences is considered.
For generating predictions we made a split of 90 \% for training and 10 \% validation to compare the performance of different models, for each specific task and based on F1 scores of a particular epoch we updated the model weights. The weights corresponding to the best validation scores have been selected for inferring the test values. We observed that usually 3, 4 trained epochs had a higher F1 score.\\
For reproducibility, the codes have been uploaded to github \footnote{https://github.com/priyanshusankhala/hasoc-hnlp}. The random seed has been set to 42.

\section{Conclusion}

In this paper, we explain the shared tasks presented by HASOC in English and Indo-Aryan languages. We used large language models which are pre-trained on large corpora for hate speech detection tasks and to evaluate predictions by different models a validation dataset was created. In future work, we hope to try out more different fine tuned models.

\begin{acknowledgments}
The authors would like to thank the organizers of Hate Speech and Offensive Content Identification in Indo-Aryan Languages 2021 \cite{hasoc2021mergeoverview} for conducting this data challenge. The authors gratefully acknowledge google colab for providing GPU’s to do the computation. All pre-trained models is based upon work supported by Hugging Face \cite{wolf2020huggingfaces}.
\end{acknowledgments}

\bibliography{w-ce}

\end{document}